\documentclass[conference]{IEEEtran}
\usepackage{times}

\usepackage{amsmath}
\usepackage{amssymb}
\usepackage{amsfonts}
\usepackage{amsthm}

\usepackage{lipsum}
\usepackage{graphicx}

\DeclareMathOperator*{\argmax}{arg\,max}

\usepackage{color}

\definecolor{dkgreen}{rgb}{0,0.6,0}
\definecolor{gray}{rgb}{0.5,0.5,0.5}
\definecolor{mauve}{rgb}{0.58,0,0.82}

\definecolor{codegreen}{rgb}{0,0.6,0}
\definecolor{codegray}{rgb}{0.5,0.5,0.5}
\definecolor{codepurple}{rgb}{0.58,0,0.82}
\definecolor{backcolour}{rgb}{0.95,0.95,0.92}

\usepackage[ruled, lined, linesnumbered, commentsnumbered, longend]{algorithm2e}

\SetCommentSty{mycommfont}

\usepackage{ifthen,version}
\newboolean{include-notes}
\setboolean{include-notes}{true}
\newcommand{\tmnote}[1]{\ifthenelse{\boolean{include-notes}}%
  {\textcolor{dkgreen}{\emph{TM says: #1}}}{}}
\newcommand{\ptnote}[1]{\ifthenelse{\boolean{include-notes}}%
  {\textcolor{red}{PT says: #1}}{}}
\newcommand{\msnote}[1]{\ifthenelse{\boolean{include-notes}}%
  {\textcolor{blue}{\emph{MS says: #1}}}{}}

\usepackage[numbers]{natbib}
\usepackage{multicol}
\usepackage[bookmarks=true]{hyperref}
\allowdisplaybreaks

\begin{document}

\title{Human Robot Pacing Mismatch}

\author{\authorblockN{Muchen Sun\authorrefmark{1},
Peter Trautman\authorrefmark{2},
and Todd Murphey\authorrefmark{1}}
\authorblockA{\authorrefmark{1}Department of Mechanical Engineering, Northwestern University, Evanston, IL 60208, USA\\muchen@u.northwestern.edu}
\authorblockA{\authorrefmark{2}Honda Research Institute, San Jose, CA 95134, USA}}

\maketitle

\begin{abstract}
A widely accepted explanation for robots planning overcautious or overaggressive trajectories alongside human is that the crowd density exceeds a threshold such that all feasible trajectories are considered unsafe---the \emph{freezing robot problem}. However, even with low crowd density, the robot's navigation performance could still drop drastically when in close proximity to human. In this work, we argue that a broader cause of suboptimal navigation performance near human is due to the robot's misjudgement for the human's willingness (flexibility) to share space with others, particularly when the robot assumes the human's flexibility holds constant during interaction, a phenomenon of what we call \emph{human robot pacing mismatch}. We show that the necessary condition for solving pacing mismatch is to model the \emph{evolution} of both the robot and the human's flexibility during decision making, a strategy called \emph{distribution space modeling}. We demonstrate the advantage of distribution space coupling through an anecdotal case study and discuss the future directions of solving human robot pacing mismatch.
\end{abstract}

\IEEEpeerreviewmaketitle

\section{Introduction}
\label{sec:introduction}

The freezing robot problem (FRP) was originally defined as follows: if ``the environment surpasses a certain level of complexity, the planner decides that all forward paths are unsafe, and the robot freezes in place (or performs unnecessary maneuvers) to avoid collisions''~\cite{trautmaniros}. Even though originally defined in a social navigation context, the FRP also impacts other human robotic interaction applications such as collaborative factory robots~\cite{shah_assembly} and assistive dressing robots~\cite{Li-RSS-21}.

This original interpretation of FRP persists in the social navigation community~\cite{unfrozen-unlost,frozone}, but only describes failures in the limit of very high crowd densities; that is, the FRP is typically thought of as a ``limit'' phenomenon, occurring \emph{only} when agent congestion is so high that the robot comes to a complete stop. However, this interpretation is incomplete: freezing robot behaviors can occur at lower densities and manifest, broadly speaking, as either overcautious (e.g. an autonomous vehicle creeping across an intersection) or overaggressive (a robot merging with pedestrian foot traffic too quickly) behavior.~\footnote{Note that neither overaggressive or overcautious behavior involve the robot actually stopping, but can still lead to highly undesirable consequences, such as slowing vehicular traffic flow or disrupting the natural cadence of pedestrian crowds.} For example, in~\cite{trautman-ijrr-2015} a 3x performance degradation is observed in the range 0.2-0.55 people/$m^2$ whereas in~\cite{trautmanicra2013} literal freezing of the robot does not occur until 0.55 people/$m^2$. 

Motivated by these observations, we seek to expand the original definition of the FRP to include the broader phenomenon of what we call \emph{human robot pacing mismatch} (HRPM). To better understand HRPM---and the methods needed to rectify this issue---we begin with an illustrative anecdote.




\begin{figure}
    \centering
    \includegraphics[width=\columnwidth]{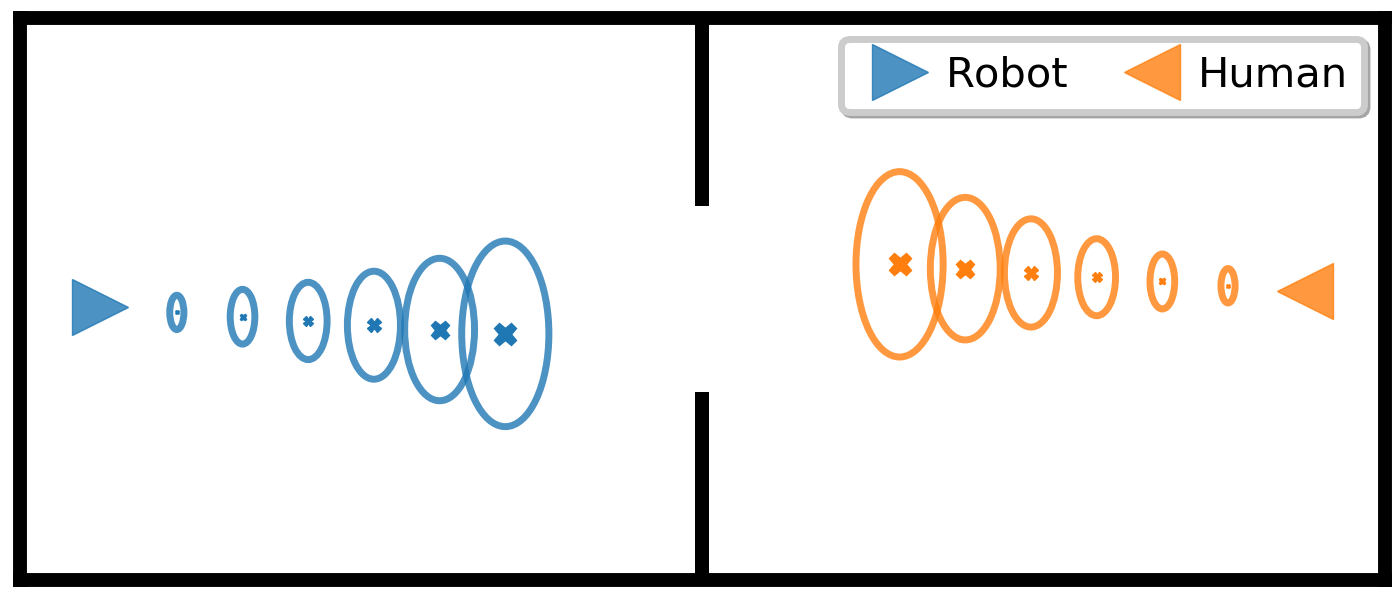}
    \caption{Example of preference distributions at each time step. The cross is the intent at each time step, while the ellipsoidal circle represents the flexibility.}
    \label{fig: definition_example}
    \vspace{-2em}
\end{figure}

\section{HRPM: Bottleneck Case Study}

\begin{figure*}
    \centering
    \includegraphics[width=\textwidth]{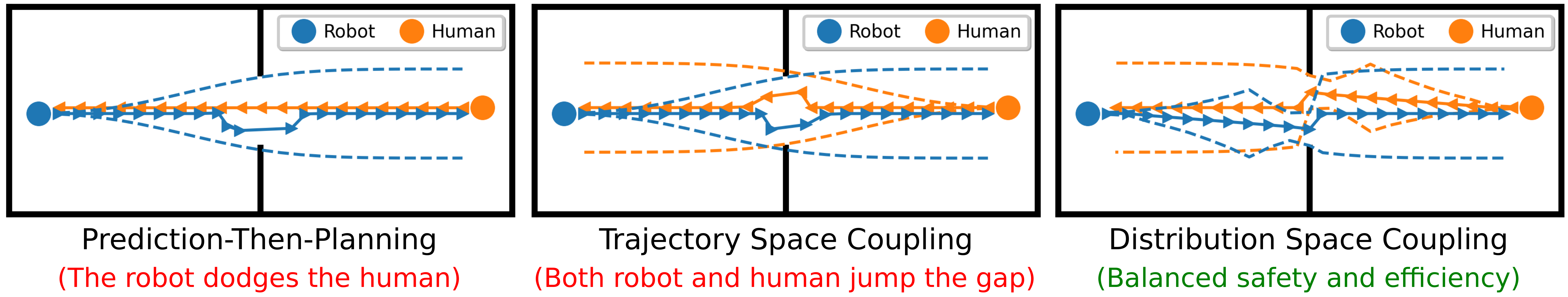}
    \vspace{-2em}
    \caption{The robot's planned trajectory and predicted pedestrian trajectory with three different strategies in the case study. The solid dots are the start position of the robot and the pedestrian. The solid lines with arrows are the predicted pedestrian trajectory and planned robot trajectory. The dashed lines form the envelope of preference distribution. {\it\textbf{Left}}: The robot predicts pedestrian trajectory ahead of planning, and does not take into account the influence from itself to the pedestrian. As a result, the robot predicts the pedestrian will leave no space for the robot to pass simultaneously, and robot chooses to dodge the human; {\it\textbf{Middle}}: The robot simultaneously predicts pedestrian trajectory and plans its own trajectory. Even though prediction and planning are coupled, the pedestrian preference is measured in open space but is used for close proximity interaction at the door. This static preference assumption leads to an incorrect estimate of pedestrian preference during interaction. As a result, the robot predicts the pedestrian's pacing is to slow down before the door and then accelerate to jump the gap, and robot plans a similarly overaggressive trajectory; {\it\textbf{Right}}: The robot predicts the evolution of pedestrian preference at each time step during interaction, while simultaneously plans the optimal preference for itself. Note that the robot predicts the pedestrian adjust the width of preference distribution near the door to leave space for the robot, and robot adjusts its preference as well in response (more details regarding the evolution of preferences during interaction can be found in Fig. \ref{fig: dso_steps}). As a result, both the robot and the pedestrian can pass the door at the same time without compromising safety or efficiency.}
    \label{fig: case_comparison}
    \vspace{-1em}
\end{figure*}


In this anecdotal case study we have one robot and one pedestrian passing through a door from different sides. 

\noindent\textbf{Assumption 1 } The door is wide enough for both the robot and the pedestrian to pass simultaneously, but the width of door is significantly smaller than the width of the room.

\noindent\textbf{Assumption 2 } There is no other obstacle, the robot and the human only need to avoid hit the wall and each other.

\noindent\textbf{Assumption 3 } The robot has a receding horizon planner, at each time step it plans a trajectory for a fixed future horizon. 

Furthermore, we define the following terms to better generalize different navigation strategies. An example of the following definitions is shown in Fig.~\ref{fig: definition_example}.

\noindent\textbf{Definition 1 (Intent) } The trajectory planned by the pedestrian and the robot without presence of obstacles is defined as the \emph{intent} of the agent. It represents the preferred and most efficient trajectory in an ideal environment. 

\noindent\textbf{Definition 2 (Preference) } Both the robot and the pedestrian evaluate the likelihood of taking a potential trajectory through the probability density function of a \emph{preference distribution}. Here we assume the preference distribution at each time step is a Gaussian distribution, with the mean being the intent at the time step. 

\noindent\textbf{Definition 3 (Flexibility) } The flexibility is defined as the agent's willingness of deviating from the intent. It is an inherent property of the preference distribution. When the preference distribution is a Gaussian distribution, we measure the flexibility as the trace of the covariance matrix---the flatter the Gaussian distribution is, the more likely a trajectory far from the intent will be taken. 


\noindent\textbf{Case: Prediction Then Planning}

In this case, the robot first predicts the pedestrian's trajectory, then plans its trajectory based on the prediction. This prediction-then-planning strategy formulates the following optimization problem:
\begin{align}
    x_{r}^*(t) & = \argmax_{x} P_{r}(x) \\
    \text{ s.t. } & S_{door}(x){+}S(x, x_{h}^*(t)) \geq \gamma \\
    & x_{h}^*(t) = \argmax_{x} P_{h}(x) 
\end{align} where $x_{\{r, h\}}(t)$ and $P_{\{r, h\}}(\cdot)$ are the robot and the pedestrian's trajectories and preference distributions, respectively. The function $S(\cdot, \cdot)$ evaluates the joint safety between two trajectories, and the function $S_{door}(\cdot)$ evaluates the safety of a trajectory with respect to the door. The constraint variable $\gamma$ is the safety threshold.

With this strategy, the robot tries to stay as close to the its intent as possible, as long as it satisfies the safety constraint. However, if the robot's prediction of pedestrian trajectory does not include the robot's influence on the pedestrian, the predicted trajectory does not necessarily leaves space for the robot. As shown in Fig. \ref{fig: case_comparison}(left), the robot predicts the pedestrian has a pace to go straight through the door without any compromise for the robot. Thus, the robot plans unnecessary maneuvers to avoid collision. whereas in reality the human is likely to leave some space for the robot to pass through simultaneously. 

\noindent\textbf{Case: Coupled Prediction and Planning of Trajectories}

A straightforward fix for the prediction-then-planning approach is to couple the prediction and planning of trajectories. By solving following joint optimization problem, the robot is able to simultaneously plan future trajectory while predicting the pedestrian's reaction to the planned trajectory\footnote{The prediction-then-planning approach can be considered as a special case of coupled prediction and planning of trajectories, where the pedestrian has infinitely small covariance matrix for the preference distribution, and thus infinitely high flexibility for taking the intent trajectory.}: 
\begin{align}
    x_{r}^*(t), x_{h}^*(t) & = \argmax_{x_{r}, x_{h}} \sum_{i\in\{r,h\}} P_i(x_i){+}S_{door}(x_i) \label{eq: tso_objective} \\
    \text{s.t. } & S(x_{r}, x_{h}) \geq \gamma
\end{align} 

In this joint optimization formalism, both the robot and the pedestrian want to stay as close to their own intent as possible, while satisfying the safety constraint. The interplay between efficiency (preserving intent) and safety (avoiding collision) is controlled by the preference distribution of each agent, in particular the flexibility of each agent. When coupling the \emph{trajectories} of two agents, the preference distributions are predicted a priori and fixed during interaction and through the whole planning horizon. 

However, an agent's (the robot or the pedestrian) preference distribution at the current time step is not necessarily the same as after interaction happens. For example, when being in the open space, the agent's preference distribution can be spread out as long as it does not overlap with the walls. But during interaction, the preference distribution should evolve in response to other agents' preferences. This is particularly true when human and robot are in close proximity, such as in our case study of navigation through a bottleneck. Under such circumstances, the assumption of static preference, even with perfect prediction at the planning time, could still lead to freezing behavior. This conclusion is validated in Fig.~\ref{fig: case_comparison}(middle), where the robot uses the prediction of pedestrian preference in an open space for close-proximity interaction at the door. As a result, the robot predicts the pedestrian will jump the gap through the door and the predicted trajectory is nearly infeasible, and robot has to plans a similarly overaggressive trajectory in response. 

\begin{figure*}[t]
    \centering
    \includegraphics[width=\textwidth]{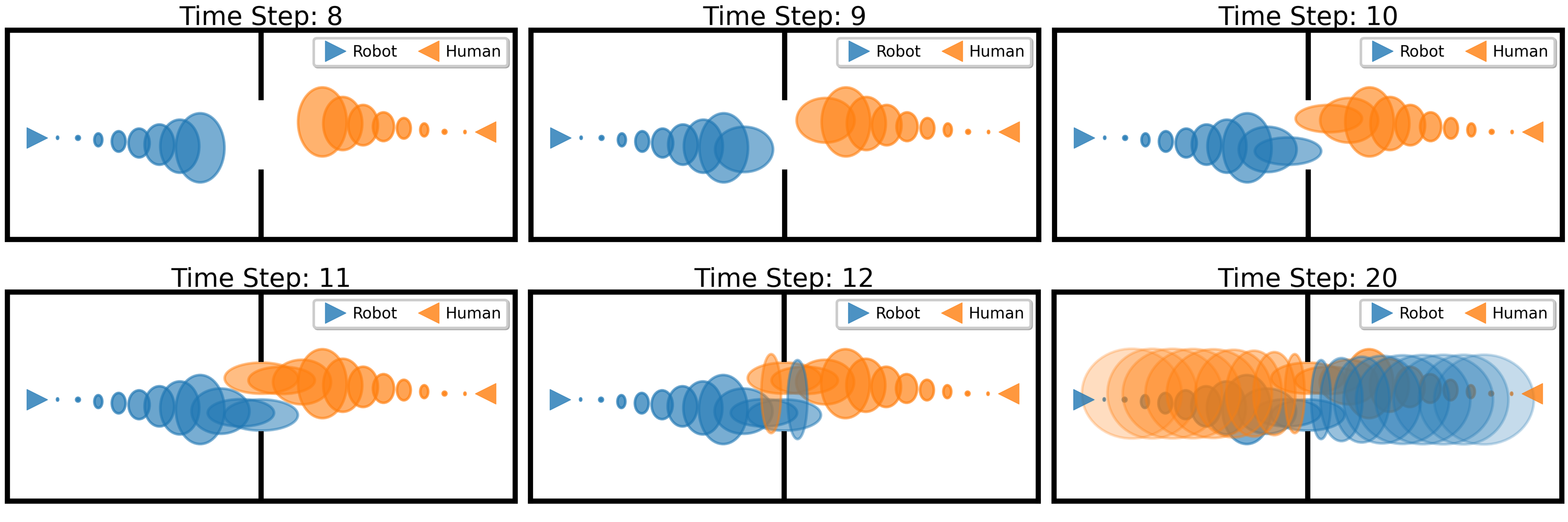}
    \vspace{-2em}
    \caption{Detailed illustration of agent preference evolution during interaction. Note that both the robot and the pedestrian adjust the flexibility near the door with the presence of close proximity interaction. An animated video of this illustration can be found at: \url{https://youtu.be/2GNeBrdHU34}.}
    \label{fig: dso_steps}
    \vspace{-1em}
\end{figure*}

\noindent\textbf{Human Robot Pacing Mismatch}


The original definition of freezing robot problem explains suboptimal behaviors, either overaggressive or overcautious, as a result of modeling choice with a paradigm---decoupled prediction and planning of trajectories. On the contrary, here we argue that human robot pacing mismatch as a broader cause of suboptimal behaviors, is from an incorrect algorithm structural assumption---the assumption that preference remains constant during interaction. In both cases above, even with a perfect prediction of prior preference, as long as the coupling of prediction and planning is within the trajectory space, suboptimal behaviors may still occur. In both cases, suboptimal behaviors are caused by the same factor: incorrect assessment of agent preference distribution during interaction, particularly due to the assumption of static preference during interaction. As a result, in this case study, the pedestrian has a walking pace for simultaneous passage through the door, while the robot incorrectly believes only one can pass at a time, and sacrifices efficiency for unnecessary safety. We call this phenomenon \emph{human robot pacing mismatch}, and formally define it as follow: 

\noindent\textbf{Definition 4 (HRPM) } Human robot pacing mismatch (HRPM) is defined as the phenomenon where the robot misjudges the human's willingness to make space for others (this willingness is often reflected by the human's pace), due to incorrect assessment of the human's preference distribution. This misjudgement eventually leads to suboptimal behaviors of the robot.


\section{Solving Human Robot Pacing Mismatch:\\Distribution Space Coupling}

\begin{figure}
    \vspace{+1em}
    \centering
    \includegraphics[width=\columnwidth]{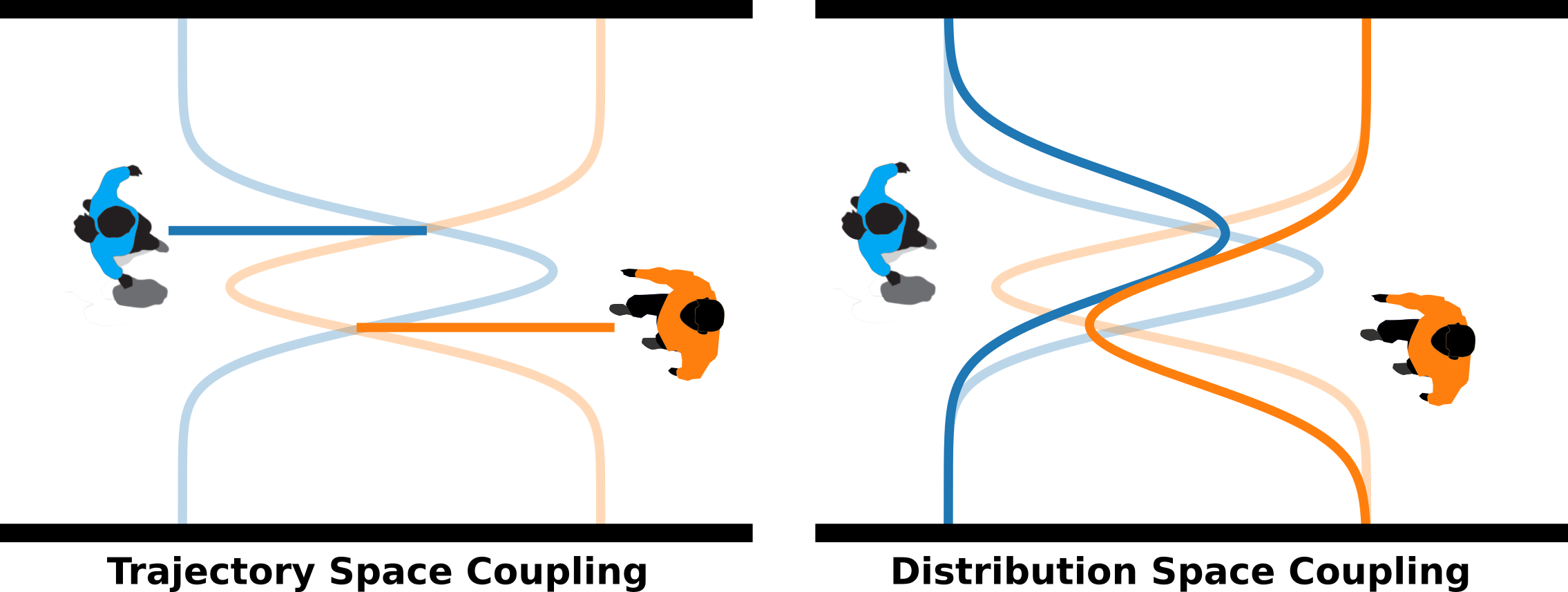}
    \vspace{-1em}
    \caption{One-dimension illustration of coupled prediction and planning in trajectory space and in distribution space. {\it\textbf{Left}}: Trajectory space coupling jointly predict and plan optimal position at each time step (solid lines), with fixed preference distributions (opaque curves); {\it\textbf{Right}}: Distribution space coupling jointly predict and plan optimal preference distributions at each time step (solid curves), which evolve from the prior preferences (opaque curves).}
    \label{fig: dso_1d}
    \vspace{-1em}
\end{figure}

As shown in the case study, even with a perfect prediction of pedestrian preference in the open space, if the prediction does not model the evolution of preference at each time step during interaction, human robot pace mismatching may still occur. Thus, one necessary condition for solving HRPM is to not plan and predict the trajectories, but instead plan and predict the preference at each time step; In other words, we need to \emph{couple prediction and planning in the space of preference distributions}---we name this approach \emph{distribution space coupling}. The general formalism of distribution space coupling is as follow:
\begin{align}
    P_{r}^*(x(t)){,}P_{h}^*(x(t)) = \argmax_{P_{r}{,}P_{h}} D_r(P_r{\Vert}\bar{P}_r){+}D_h(P_h{\Vert}\bar{P}_h)  \\
    \text{s.t. } \mathbb{E}_{P_{r},P_{h}}[S(\cdot,\cdot)] + \mathbb{E}_{P_r}[S_{door}(\cdot)] + \mathbb{E}_{P_h}[S_{door}(\cdot)] \geq \gamma
\end{align} where 
\begin{align}
    \mathbb{E}_{P_{r},P_{h}}[S(\cdot,\cdot)] & = \int\int S(x_r, x_h) P_{r}(x_r) P_{h}(x_h) dx_r dx_h \label{eq: expected_jiont_safety} \\
    \mathbb{E}_{P_i}[S_{door}(\cdot)] & = \int S_{door}(x) P_i(x) dx , \quad i\in\{r,h\}
\end{align} and $D_i(P_i \Vert \bar{P}_i)$ measures the ``similarity'' between two distributions $P_i(x)$ and $\bar{P}_i(x)$, with $\bar{P}_i(x)$ being the prior preference of the agent. A one-dimensional illustration in shown in Fig.~\ref{fig: dso_1d}.

\noindent\textbf{Remark 1 } In trajectory space coupling, the collision avoidance function $S(\cdot,\cdot)$ serves as the ``safety zone''---a function specifies how comfortable agents feel about being close to others. Therefore, it must has a non-finite support. However, any collision avoidance function with non-zero support imposes an unwarranted prior on top of agents' preferences. Furthermore, in the trajectory space coupling objective function (\ref{eq: tso_objective}), the collision avoidance function is shared by the agents, but agents do not necessarily feel same about being close to each other. On the other hand, distribution space coupling can fully preserve the statistics of agent preferences by using the Dirac delta function as the collision avoidance function. The Dirac delta function has a support of $\{0\}$, thus it imposes the least possible prior structure on the collision avoidance function. It can also simplify the expected safety between preference distributions (\ref{eq: expected_jiont_safety}) as:
\begin{align}
    \mathbb{E}_{P_{r},P_{h}}[S(\cdot,\cdot)] & = \int\int \delta(\vert x_r{-}x_h \vert) P_{r}(x_r) P_{h}(x_h) dx_r dx_h \nonumber \\
    & = \int P_{r}(x_r) P_{h}(x_h) dx_r dx_h \nonumber \\
    & = \langle P_{r},P_{h}\rangle
\end{align} where $\langle\cdot,\cdot\rangle$ is the inner product in function space. Essentially it means the preference distribution itself serves as the ``safety zone''. 

\noindent\textbf{Case: Gaussian Space Coupling}

As a validation of distribution space coupling, we implement a coupled prediction and planning method in the space of Gaussian distributions, with the same setup from the bottleneck case study. We constrain the preference distribution at every time step to be a two-dimensional Gaussian distribution, and we model the evolution of both the mean and covariance at each time step during interaction. We use Dirac delta function as the collision avoidance function and Kullback-Leibler divergence to measure the similarity between two distributions. As shown in Fig. \ref{fig: case_comparison}(right), by coupling preference distributions between agents, the robot generates a more natural prediction of human trajectory and plans a similarly natural trajectory for itself. It is worth noting that, with distribution space coupling both agents start to adjust trajectory (intent) further from the door compared with trajectory space coupling and prediction-then-planning. It is also worth noting that the smoothness of the trajectories is partially the benefit of co-evolution of intent and flexibility---both agents avoid unnecessary maneuvers by allocating part of safety gain through adjusting flexibility.

\section{Conclusion and Future Directions}

In this work we introduce a broader phenomenon for causing suboptimal navigation behavior when robots are near human---\emph{human robot pacing mismatch}(HRPM), a phenomenon where the robot exhibits suboptimal behavior due to misjudgement of the human's willingness to make space for others. We argue that a necessary condition to solve HRPM is to model the evolution of agent preference during interaction, and we name this approach as \emph{distribution space coupling}. We demonstrate the advantages of distribution space coupling through an anecdotal case study. 

However, there remains a gap between distribution space coupling and practical social navigation algorithms. Below we address several challenges and future directions. 

\noindent\textbf{Higher order statistics of preference distribution}

While the Gaussian preference assumption represents flexibility with the covariance matrix, but there are higher order statistical feature of the preference distribution, such as skew and kurtosis, that reflect flexibility as well. In \cite{SunM-RSS-21}, a distribution space coupling algorithm is proposed without the Gaussian assumption, however it relies on samples to represent the underlying preference, it is still challenging to analyze higher order statistics of the evolving preference distributions.

\noindent\textbf{Representation of preference distribution}

It is necessary to look beyond the Gaussian representation of preference distribution, since the preference could be asymmetric and multi-modal. The method in \cite{SunM-RSS-21} could capture arbitrary distribution but its practical performance is constrained by sampling efficiency.

\noindent\textbf{Efficient optimization in distribution space}
Distribution space coupling offers an unique opportunity where a global minimum in distribution space can be a nonconvex distribution~\cite{SunM-RSS-21}. However, efficiency optimization methods are still in need for real-time social navigation.





\bibliographystyle{plainnat}
\bibliography{bibliography}

\end{document}